\documentclass[review]{elsarticle}
\usepackage[utf8]{inputenc}
\usepackage{amsmath,amsthm, amssymb}
\usepackage{bm}
\usepackage{graphicx}
\usepackage{array}
\usepackage{float}
\usepackage{algorithmic,algorithm}

\usepackage{subfigure}
\usepackage{multirow}
\usepackage[figuresright]{rotating}
\usepackage{booktabs}

\newtheorem{MyThe}{Theorem}
\newtheorem{MyCor}{Corollary}
\newtheorem{MyLem}{Lemma}
\begin{document}
\begin{frontmatter}
\title{$\bar{G}_{mst}$:An Unbiased Stratified Statistic and a Fast Gradient Optimization Algorithm Based on It.
\tnoteref{mytitlenote}}
\tnotetext[mytitlenote]{Research supported by Chinese National Office for Philosophy and Social Sciences(19FTJB003),National Bureau of Statistics of China($2016LY64$), Natural Science Foundation of Guangdong Province (10451032001006140,2015A030313623).}
\author[mymainaddress]{Aixiang(Andy) Chen}

\address[mymainaddress]{Institute of Artificial Intelligence and Deep Learning, Guangdong University of Finance and Economics,  \small Guangzhou 510320,China}

\begin{abstract}
The fluctuation effect of gradient expectation and variance caused by parameter update between consecutive iterations is neglected or confusing by current mainstream gradient optimization algorithms. The work in this paper remedy this issue by introducing a novel unbiased  stratified statistic \ $\bar{G}_{mst}$\ , a sufficient condition of fast convergence for \ $\bar{G}_{mst}$\  also is established.  A novel algorithm named MSSG designed based on  \ $\bar{G}_{mst}$\  outperforms other sgd-like algorithms. Theoretical conclusions and experimental evidence strongly suggest to employ MSSG when training deep model.
\end{abstract}


\end{frontmatter}
\section{Introduction}\label{intro}
In recent years, by optimizing a deep and big model, machines have achieved capabilities comparable to or even surpassing humans in many fields such as image processing and visual recognition\cite{andy2020,LeCun1998-3,Ciresan2010,Ciresan2012,KrizhevskySH12-2}, speech recognition\cite{Graves2006,Graves2013-4,Shillingford2018}, machine translation and natural language understanding\cite{Yonghui2016-4}. The gradient method (SGD-like Algorithms) developed based on the work of Robbins and Monro\cite{Robbins1951} is still the mainstream and competitive algorithm for training deep models.

It is difficult to optimize a giant model with deep and wider layers. Similar to most optimization algorithms, training a deep model with gradient method (SGD-like Algorithms) has disadvantages such as easy to fall into local minima or saddle point and slow convergence speed. There have been a lot of researches on the improvement of the gradient method, and a considerable part of these researches focus on how to refine the search direction while keeping the iteration cost as low as possible to accelerate the convergence of the algorithm\cite{Robbins1951,Qian1999,Nesterov,Roux2012,Defazio2014,Johnson2013,Andy2018}.

These improvements for the search direction are roughly divided into two categories. One is the momentum method\cite{Qian1999} based on the principles of physics and the corresponding improved algorithms\cite{Nesterov,Kingma2015,Dozat2016}, the momentum method avoids excessive swing amplitude of the search track by retaining part of the potential energy of the original track to accelerate the convergence.

The other is various algorithms based on the idea of variance reduction\cite{Polyak1992,Nesterov2009,Roux2012,Defazio2014,Johnson2013,Andy2018}. In the variance reduction method, except batch stochastic gradient and full gradient, other methods involve the use of historical gradient information to reduce the orbital variance, which is essentially the same as the momentum method.

Whether it is the momentum method or the variance reduction method, the theoretical research is still insufficient. For example, although literature \cite{Qian1999} discusses the relationship between the momentum coefficient \ $p$\  and the rate of convergence in detail, how to determine the value of the momentum coefficient \ $p$\  to ensure that the parameter update direction is an unbiased estimator of the global minimum of the cost function. There is no clear Theoretical results. For another example, variance reduction by averaging is a commonly used strategy, but how to set the range of averaging (or the window width for averaging)appropriately, the existing research has not given a clear answer to this. The consequence of these deficiencies is that the specific effects of the improvement strategy rely on experimental observations on a limited data set, and it is difficult to achieve universal and convincing improvement effects in practice.

This paper proposes a new strategy called memory-type stochastic stratification gradient. The\ $\bar{G}_{mst}$\ designed around this strategy can be regarded as the general case of momentum method and variance reduction method. Based on the\ $\bar{G}_{mst}$\ statistic, we establishes sufficient conditions to make \ $\bar{G}_{mst}$\  unbiased and its variance to decay quickly, which provides a theoretical basis for the setting of the momentum coefficient and the range of average.

This paper tests the results on several random artificial datasets and benchmark dataset MNIST. The experimental results show that \ $\bar{G}_{mst}$\  can provide a more stable and accurate search direction, and improve the test accuracy of the model, thereby improving the generalization ability of the model.

\section{$\bar{G}_{mst}$\ Statistics}

Generally, given training data set $(X,Y)$, the object function to be optimized for a deep model is a finite sum of loss functions $J_i$ as follows.
\begin{equation}\label{eq:cost}
\arg\min_{W\in R^p}J(W,b)=\frac{1}{N}\sum_{i=1}^{N}J(W,b;x^{(i)},y^{(i)})=\frac{1}{N}\sum_{i=1}^{N}J_{i}
\end{equation}
where $(W,b)$ are optimized parameters, $J_i=J(W,b;x_i,y_i)$ is the loss function on the $j^{th}$ sample in $(X,Y)$, the number $N$ is the size of the training data. For example, when the optimized model is a three layers of neural network with $S_1$ input nodes, $S_2$ hidden nodes and $S_3$ output nodes, the model can be expressed in the form of $W\in R^{S_1\times S_2\times S_3}$.

Similar to the SGD-like Algorithms, \ $\bar{G}_{mst}$\  in this article is a statistic designed to optimize the cost function in the form of formulae \ \ref{eq:cost}\ , and the algorithm MSSG(algorpithm \ \ref{alg:barGmstzeromean}\ ) designed around \ $\bar{G}_{mst}$\  is  well-suited for solving a classification problem with\ $C$\ (\ $C\geq2$\ )categories. MSSG uses iterations of the form to perform parameter update:
\begin{equation}\label{eq:weightupdate}
W^{k+1}=W^{k}-h\bar{G}^{k}_{mst}=W^{k}-h\sum\limits_{j=1}^{C}\textit{w}_j\cdot G_{j}^{k}
\end{equation}
where the statistics $\bar{G}_{mst}$\ is the mean of a $C$-dimensions vector $G$ by component, the superscripts on the variables are used to denote the index of iteration,the $h$ is step size of learning algorithm, the italic $\textit{w}_j$ is weight for class $j=c(1\leq c\leq C)$,determined by the ratio of sample to total in number.

The auxiliary vector $G$ in formulae \ref{eq:weightupdate} tracks the gradient signal ever used, this is the origin of $MSSAG$ being renamed the memory algorithm.  the $k^{th}$ iteration a random index\ $\xi_j$\ for each $j=c\in 1\leq c\leq C$\ is selected and we set
\begin{equation}\label{eq:Gvector}
G_{j}^{k}=
\begin{aligned}
p_j^k\cdot G_{j}^{k-1}+q_j^k\cdot g(W^k,\xi_j) &\ for\ each\ \substack{j\in \{1,\cdots,C\},\\k\in \{1,2,\cdots,\}}
\end{aligned}
\end{equation}

where $\xi_j$ is a random index of sample in category $j=c_{k}$,the $g(W^k,\xi_j)$ is random gradient generated by network $W^k$ inputting this sample.

If we know the value of \ $p_j,q_j$\ in formulae \ref{eq:weightupdate}\ , MSSG algorithm based on \ $\bar{G}_{mst}^k$\ produce a gradient sequence \ $\bar{G}_{mst}^{k}(k=1,2,\cdots,)$\ as iteration proceeds. Inappropriate values of \ $p_j,q_j$\  will destroy the unbiasedness of \ $\bar{G}_{mst}^k$\ . The conditions for making \ $\bar{G}_{mst}^k$\  unbiased estimation of the overall gradient mean \ $\bar{G}^{k}$\  will be discussed below.

\subsection{\ $p,q$\ conditions to ensure \ $\bar{G}_{mst}^k$\ an unbiased estimation}
The basic idea to ensure  \ $\bar{G}_{mst}^k$\  an unbiased estimation is to convert the gradient signal of the previous iteration to the current iteration in an appropriate proportion. This result is given in the form of theorem below.

\begin{MyThe}\label{the:unbias}
$G(W^{k-1},\xi_j),G(W^{k},\xi_j)(j=1,\cdots,C)$\  denote random gradients of network\ $W^{k-1},W^{k}$\  when inputting a random sample  \ $\xi_j$\  from \ $j^{th}$\ class(that is, the random gradients are produced in two consecutive iterations),\ $E^{k-1}_j,E^k_j$\ are their expectations respectively. If\ $\frac{p_j^k}{1-q_j^k}=\frac{E^k_j}{E^{k-1}_j}$\ , then\ $\bar{G}_{mst}^k$\ is an unbiased estimation of population mean \ $\bar{G}$\ , that is \ $E(\bar{G}_{mst}^k)=\bar{G}^k$.
\end{MyThe}

\begin{proof}

In order to ensure that the starting point \ $\bar{G}_{mst}^1$\  of the sequence \ $\{\bar{G}_{mst}^k\}$\  is also an unbiased estimate of \ $\bar{G}^{1}$\ , without loss of generality, we randomly select a sample from each category $j\in \{1,2,\cdots C\}$ in advance, calculate its gradient, fill the \ $G$\  vector and calculate \ $\bar{G}_{mst}^1$\  accordingly as the starting point of the sequence \ $\{\bar{G}_{mst}^k\}$\ . Therefore,at starting point, we have \ $\bar{G}_{mst}^1=\bar{G}_{st}^1=\sum\limits_{j=1}^C\textit{w}_jG_j^1$\ .

Taking an expectation of \ $\bar{G}_{mst}^1$\  will leads to \ $E(\bar{G}_{mst}^1)=E(\bar{G}_{st}^1)=E(\sum\limits_{j=1}^C\textit{w}_jG_j^1)=\sum\limits_{j=1}^C\textit{w}_jE(G_j^1)=\bar{G}^1$\ , The unbiasedness of the starting point of the gradient sequence is established.

Similarly, for  \ $k^{th}$\ iteration, we have:

\begin{equation}
\begin{array}{rl}
E(\bar{G}_{mst}^k)=&E(\sum\limits_{j=1}^C\textit{w}_jG_j^k)=\sum\limits_{j=1}^C\textit{w}_jE(G_j^k)\\
=&\sum\limits_{j=1}^C\textit{w}_jE[p_j^kG_j^{k-1}+q_j^kG(W^k,\xi_j)]\\
=&\sum\limits_{j=1}^C\textit{w}_j[p_j^kE(G_j^{k-1})+q_j^kE(G(W^k,\xi_j))]\\
=&\sum\limits_{j=1}^C\textit{w}_j[p_j^kE_j^{k-1}+q_j^kE_j^k]\\
=&\sum\limits_{j=1}^C\textit{w}_j[(1-q_j^k)E_j^k+q_j^kE_j^k]\\
=&\sum\limits_{j=1}^C\textit{w}_jE_j^k=\bar{G}^k
\nonumber
\end{array}
\end{equation}
This conclude the proof.
\end{proof}

\subsection{Discussion on variance of  unbiased statistics \ $\bar{G}_{mst}^k$\  }
In this section, the following lemma is first given below. We then discuss the properties of the variance of \ $\bar{G}_{mst}^k$\ on the basis of this lemma.

\begin{MyLem}\label{the:varofbarGmst}
Let \ $E_j^{k-1},E_j^k$\ be the mean of \ $j^{th}$\ sub-population of network\ $W^{k-1},W^k$\  , \ $V_j^{k-1}=V(G_j^{k-1}),V_j^k=V(G_j^{k})$\ be the variance of \ $j^{th}$\ sub-population of network\ $W^{k-1},W^k$\  respectively. the variance of \ $\bar{G}_{mst}^k$\ is given by:
\begin{equation}\label{eq:Vmin}
V_{sp}(\bar{G}_{mst}^k)=\sum\limits_j^C[\textit{w}_j^2\frac{(E_j^k)^2V_j^{k-1}\cdot V_j^{k}}{(E_j^k)^2V_j^{k-1}+(E_j^{k-1})^2V_j^{k}}]
\end{equation}
when
\begin{equation}\label{eq:valuepq}
p_j^k=\frac{E_j^kE_j^{k-1}V_j^k}{(E_j^k)^2V_j^{k-1}+(E_j^{k-1})^2V_j^k},q_j^k=\frac{(E_j^k)^2V_j^{k-1}}{(E_j^k)^2V_j^{k-1}+(E_j^{k-1})^2V_j^{k}}
\end{equation}
\end{MyLem}
\begin{proof}
We know\ $\bar{G}_{mst}^k=\sum\limits_j^C\textit{w}_jG_j^k=\sum\limits_j^C[\textit{w}_j(p_j^kG_j^{k-1}+q_j^kG(W^k,\xi))]$\ , the variance of \ $\bar{G}_{mst}^k$\ is:
\begin{equation}\label{eq:var}
\begin{array}{rl}
V(\bar{G}_{mst}^k)=&\sum\limits_j^C[\textit{w}_j^2((p_j^k)^2V_j^{k-1}+(q_j^k)^2V(G(W^k,\xi)))]\\
=&\sum\limits_j^C[\textit{w}_j^2((p_j^k)^2V_j^{k-1}+(q_j^k)^2V_j^k)]
\end{array}
\end{equation}
Since \ $\bar{G}_{mst}^k$\ is an unbiased estimator, Substituting the unbiased condition \ $\frac{p_j}{1-q_j }=\frac{E_j^k}{E_j^{k-1}}$\ in theorem \ref{the:unbias}\ into formulae\ \ref{eq:var}\  leads to:

\begin{equation}\label{eq:var2}
\begin{array}{rl}
z_j^k=&(1-q_j^k)^2(\frac{E_j^k}{E_j^{k-1}})^2V_j^{k-1}+(q_j^k)^2V_j^k
\end{array}
\end{equation}
Obviously, when \ $p_j^k,q_j^k$\ take the values according to formulae\ \ref{eq:valuepq}\ , $z_j^k$ in formulae\ \ref{eq:var2}\ equal\ $\frac{(E_j^k)^2V_j^{k-1}\cdot V_j^{k}}{(E_j^k)^2V_j^{k-1}+(E_j^{k-1})^2V_j^{k}}$\ ,Substituting it back into Equation \ \ref{eq:var}\  can get the final variance expression in the form of Equation \ \ref{eq:Vmin}\ . We conclude the proof.
\end{proof}

According to Equation \ \ref{eq:valuepq}\ , \ $q_j^k$\ can be guaranteed to be less than \ $1$\ due to \ $V_j^k\neq 0$\ . However, the parameter\ $p_j^k$\ may be great than\ $1$\  without additional restrictions, the following theorem \ref{The:p-less-one}\ gives a sufficient condition to make \ $p_j^k<1$\ .

\subsection{Design effect of statistics \ $\bar{G}_{mst}^k$}
The aforementioned general results are not easy to see the effect of the new statistic \ $\bar{G}_{mst}^k$\  on variance reduction. In fact, the variance of the layer (category) samples is left in each component of the vector of memory \ $G$\  by \ $\bar{G}_{mst}^k$\  in different proportions \ $p_j^k(j\in\{1,\cdots,C\},k\in\{1,2,\cdots,\})$\ , and is rapidly attenuated as the iteration proceeds. Therefore, the variance of the statistic \ $\bar{G}_{mst}^k$\ (memory type) is smaller than that of the traditional stratified sampling statistic \ $\bar{G}_{st}^k$\  (memoryless type), and the variance remaining in the memory part will be rapidly attenuated as the iteration progresses.

\begin{MyCor}\label{cor:desigeneffect}
\ $\bar{G}_{mst}^k$\ have the following properties:
\begin{enumerate}[(1)]
\item $V_{sp}(\bar{G}_{mst}^k)<V(\bar{G}_{st}^k)$
\item $V_{sp}(\bar{G}_{mst}^{k+t})\leq p^{2t}V(\bar{G}_{mst}^{k})+\sum\limits_{i=1}^{t}p^{2(t-i)}q^2V(\bar{G}_{st}^{k+i}),0<p,q<1$
\end{enumerate}
\end{MyCor}
\begin{proof}

Property (1) is easy to deduce from the equivalent transformation of Equation \ \ref{eq:Vmin}\ . Dividing the numerator and denominator of the fraction in Formula \ \ref{eq:Vmin}\  by \ $(E_j^{k-1})^2V_j^{k-1}$\ , set \ $\rho_j=\frac{(E_j^k/E_j^{k-1})^2}{(E_j^k/E_j^{k-1})^2+V_j^k/V_j^{k-1}}<1$\ , and take \ $\rho=\max\limits_j\rho_j,j\in\{1,\cdots,C\}$\ , then the following inequality holds:
\begin{equation}
\begin{array}{rl}
V_{sp}(\bar{G}_{mst}^k)=&\sum\limits_j^C[\textit{w}_j^2\frac{(E_j^k)^2V_j^{k-1}\cdot V_j^{k}}{(E_j^k)^2V_j^{k-1}+(E_j^{k-1})^2V_j^{k}}]\\
=&\sum\limits_j^C[\textit{w}_j^2\frac{(E_j^k/E_j^{k-1})^2\cdot V_j^{k}}{(E_j^k/E_j^{k-1})^2+V_j^{k}/V_j^{k-1}}]\\
=&\sum\limits_j^C[\textit{w}_j^2\rho_jV_j^k]\\
<&\rho\sum\limits_j^C[\textit{w}_j^2V_j^k]=\rho V(\bar{G}_{st}^k)\\
<&V(\bar{G}_{st}^k) \nonumber
\end{array}
\end{equation}
Therefore, property (1) holds.

The following is a  proof of property (2) by induction. First, assume that the maximum values of all possible \ $p_j^k,q_j^k$\  generated in the iteration are \ $p,q$\  respectively, that is, \ $p=\max\limits_{\substack{j\in\{1,\cdots,C\},\\ k\in\{1,2,\cdots,\}}}p_j^k,q=\max\limits_{\substack{j\in\{1,\cdots,C\},\\ k\in\{1,2,\cdots,\}}}q_j^k$\ . When \ $k=1$\ , \ $\bar{G}_{mst}^1=\bar{G}_{st}^1$\  is known according to the aforementioned Theorem \ \ref{the:unbias}\ . According to formula\ \ref{eq:weightupdate}\ ,we know:
\begin{equation}\label{eq:bargmst2}
\begin{array}{rl}
\bar{G}_{mst}^2=&\sum\limits_{j=1}^C\textit{w}_j(p_j^2\cdot G_{j}^{1}+q_j^2\cdot g(W^2,\xi_j))\\
=&\sum\limits_{j=1}^C\textit{w}_j(p_j^2\cdot G_{j}^{1})+\sum\limits_{j=1}^C\textit{w}_j(q_j^2\cdot g(W^2,\xi_j))\\
\leq &p\cdot\sum\limits_{j=1}^C\textit{w}_jG_{j}^{1}+q\cdot\sum\limits_{j=1}^C\textit{w}_jg(W^2,\xi_j)\\
=&p\cdot \bar{G}_{mst}^1+q\cdot \bar{G}_{st}^2
\end{array}
\end{equation}

Taking the variance on both sides of inequality \ \ref{eq:bargmst2}\ , we can get \ $V_{sp}(\bar{G}_{mst}^2)\leq p^2V(\bar{G}_{mst}^1)+q^2V(\bar{G}_{st}^2)$\ , which is the situation when \ $k=1,t=1$\  in property (2). Therefore, the inequality in property (2) holds when \ $k=1,t=1$\ .

It is not difficult to verify that the inequality \ $V(\bar{G}_{mst}^{k+1})\leq p^2V(\bar{G}_{mst}^k)+q^2V(\bar{G}_{st}^{k+1})$\  is also valid, that is, only when \ $t=1$\   and $k$\ takes any value, the inequality in property (2) is valid.
\begin{equation}\label{eq:bargmstk}
\begin{array}{rl}
V_{sp}(\bar{G}_{mst}^{k+n+1})\leq&p^2\cdot V(\bar{G}_{mst}^{k+n})+q^2\cdot V(\bar{G}_{st}^{k+n+1})\\
\leq&p^2\cdot [p^{2n}V(\bar{G}_{mst}^{k})+\sum\limits_{i=1}^np^{2(n-i)}q^2V(\bar{G}_{st}^{k+i})]+q^2\cdot V(\bar{G}_{st}^{k+n+1})\\
=&p^{2(n+1)}V(\bar{G}_{mst}^{k})+p^2\sum\limits_{i=1}^np^{2(n-i)}q^2V(\bar{G}_{st}^{k+i})+q^2\cdot V(\bar{G}_{st}^{k+n+1})\\
=&p^{2(n+1)}V(\bar{G}_{mst}^{k})+\sum\limits_{i=1}^{n+1}p^{2(n+1-i)}q^2V(\bar{G}_{st}^{k+i})
\end{array}
\end{equation}

Assuming that \ $t=n$\ and $k$\ takes an arbitrary value, the inequality in property (2) holds. The derivation process of Equation \ \ref{eq:bargmstk}\  shows that when \ $t=n+1$\  and \ $k$\  takes any value, the inequality in property (2) still holds.

In summary, the inequality in property (2) holds for any k and t. The proof is complete.
\end{proof}

The property (1) in Corollary \ \ref{cor:desigeneffect}\  shows that the memory-type statistic \ $\bar{G}_{mst}^k$\  has a smaller design effect (smaller variance) than the stratified sampling statistic \ $\bar{G}_{st}^k$\

The property (2) in Corollary \ \ref{cor:desigeneffect}\  shows that the expansion or contraction of \ $V_{sp}(\bar{G}_{mst}^{k+t})$\  mainly depends on the value of the parameter \ $p$\ . The following Theorem \ \ref{The:p-less-one}\  gives sufficient conditions to ensure that \ $V_{sp}(\bar{G}_{mst}^{k+t})$\  decays rapidly with the number of iteration steps.

\begin{MyThe}\label{The:p-less-one}
Let \ $p=\max\limits_{\substack{j\in\{1,\cdots,C\},\\ k\in\{1,2,\cdots,\}}}p_j^k$\ ,where \ $p_j^k$\  is given by equation \ \ref{eq:valuepq}\ . if \ $E_j^k=E_j^{k-1}$\ , then \ $V_{sp}(\bar{G}_{mst}^{k+t})$\  decay at the rate of \ $p^{2t}$\
\end{MyThe}
\begin{proof}

According to Equation \ \ref{eq:valuepq}\ , when \ $E_j^k=E_j^{k-1}$\ , under the fact that \ $(E_j^k)^2V_j^{k-1}\geq 0$\ , \ $p_j^k\leq 1$\ , so \ $p\leq 1$\ .

Furthermore, Let \ $\frac{0}{0}=1$\ , when \ $E_j^k=E_j^{k-1}=0,V_j^k\neq 0$\ , \ $p_j^k=\frac{\frac{E_j^k}{E_j^{k-1}}V_j^k}{(\frac{E_j^k}{E_j^{k-1}})^2V_j^{k-1}+V_j^k}=\frac{V_j^k}{V_j^{k-1}+V_j^k}<1$\ , so \ $p<1$\ .

Under the condition of \ $p<1$\ , according to property (2) in Corollary \ \ref{cor:desigeneffect}\ , the variance of \ $V_{sp}(\bar{G}_{mst}^{k+t})$\  remaining in the memory will decay at a rate of \ $p^{2t}$\ .
\end{proof}

\section{MSSG algorithm based on \ $\bar{G}_{mst}$\ }

The update direction of MSSG is determined by calculating the mean value of $G$ by component, which means it needs to maintain a $C$-dimensions vector $G$ during iterations. At each iteration MSSG calculates a mini-batch gradient mean $G_j$ of samples of the $j^{th}$ class, and then the $j^{th}$ component in $G$ is updated by the new $G_j$ for each class $j^{th}\in \{0,1,\cdots C \}$.

The implementation pseudo code of MSSG is described in algorithm \ref{alg:barGmstzeromean}, which is designed according to Theorem\ \ \ref{The:p-less-one}\  to ensure convergence.
\begin{algorithm}
\caption{Memory-type Stochastic Stratified Gradient(MSSG) for minimizing $\frac{1}{N}\sum_{i=1}^{N}J_i(W)$ with step size h}
\label{alg:barGmstzeromean}
\begin{algorithmic}[1]
\STATE Parameters:Step size $h$, Batch size $B$ ,Training data size $N$, the total number of class $C$, the $j^{th}$ class size $N_j$
\STATE Inputs:training data $\small{(x^{(1)},y^{(1)}),(x^{(2)},y^{(2)}),\cdots,(x^{(N)},y^{(N)})}$
\STATE set $G_j=0$ for $j=c\in\{1,2,\cdots,C\}$,cnt=0
\WHILE{ cnt$\le$iter\_num }
\FOR{$c=0,1,\cdots C $ }
\STATE calculate sample mean $E_j$ and sample variance $V_j$ of $\frac{B}{N}$ samples from the $j^{th}$ class data
\STATE calculate $p_j,q_j$ according to equation \ref{eq:valuepq}
\STATE update $j^{th}$ component of G,$G_j\leftarrow p_j\times G_j+q_j\times (E_j-g(W^k,x_{\xi}))$, where $x_{\xi}$ is a random sample from $j^{th}$ class
\ENDFOR
\STATE $W^{k+1}=W^k-\frac{h}{C}\cdot \sum\limits_{j=1}^{C}\frac{N_j}{N}\times (G_j+E_j)$
\STATE cnt+=1
\ENDWHILE
\end{algorithmic}
\end{algorithm}

\section{Experimental results}\label{sec:experimence}
In order to verify that \ $\bar{G}_{mst}$\  has a higher estimation accuracy than that of \ $\bar{G}_{st}$\ ,\ $SGD$\ , and\ $Batch$\ ,  a form of \ $40\times 10$\  random matrix(that is 40 random numbers in each round,10 rounds successively ) be generated as an artificial data set for testing the performance of different algorithms. The artificial data is divided into four categories:increasing mean, decreasing mean, increasing variance, and decreasing variance.  For each round of 40 data, we divide them into 4 sub-populations in the order of \ $1\sim 10,11\sim 20,21\sim 30,31\sim 40$\ , as a layered simulation. In the experiment, we focus on the square of the deviation between the estimated value generated by the estimator and the overall mean (true value) to evaluate the accuracy of the search direction provided by the estimator.

Based on fair and comparable considerations,the number of samples for all methods is \ $4$\ , except for \ $SGD$\  which uses a single sample.\ $\bar{G}_{mst},\bar{G}_{st}$\  randomly select a sample from each of the \ $4$\  subpopulations, and \ $Batch$\  randomly select \ $4$\  samples from \ $40$\  populations.

The parameters \ $p_j^k,q_j^k$\  in \ $\bar{G}_{mst}$\  are calculated according to formulae \ \ref{eq:valuepq}\ , and the \ $E_j,V_j$\  involved in formulae \ \ref{eq:valuepq}\  are respectively substituted with the mean and variance of the sub-populations in the random data set for calculation.

\subsection{Results on a uniformly random data set}
The first  data uses a set of uniformly distributed random numbers from \ $10$\  intervals \ $[8,12],[8,10],[6,9],[5,8],[4,7],[3,6],[3,5],[2,4],[2,3],[0,3]$\  (decreasing mean) respectively,to form a random data set with a specification of \ $40\times 10$\  as the data population.

\begin{figure}[htbp]
  \begin{minipage}[t]{\linewidth}
  \centering
    \includegraphics[width=0.8\textwidth]{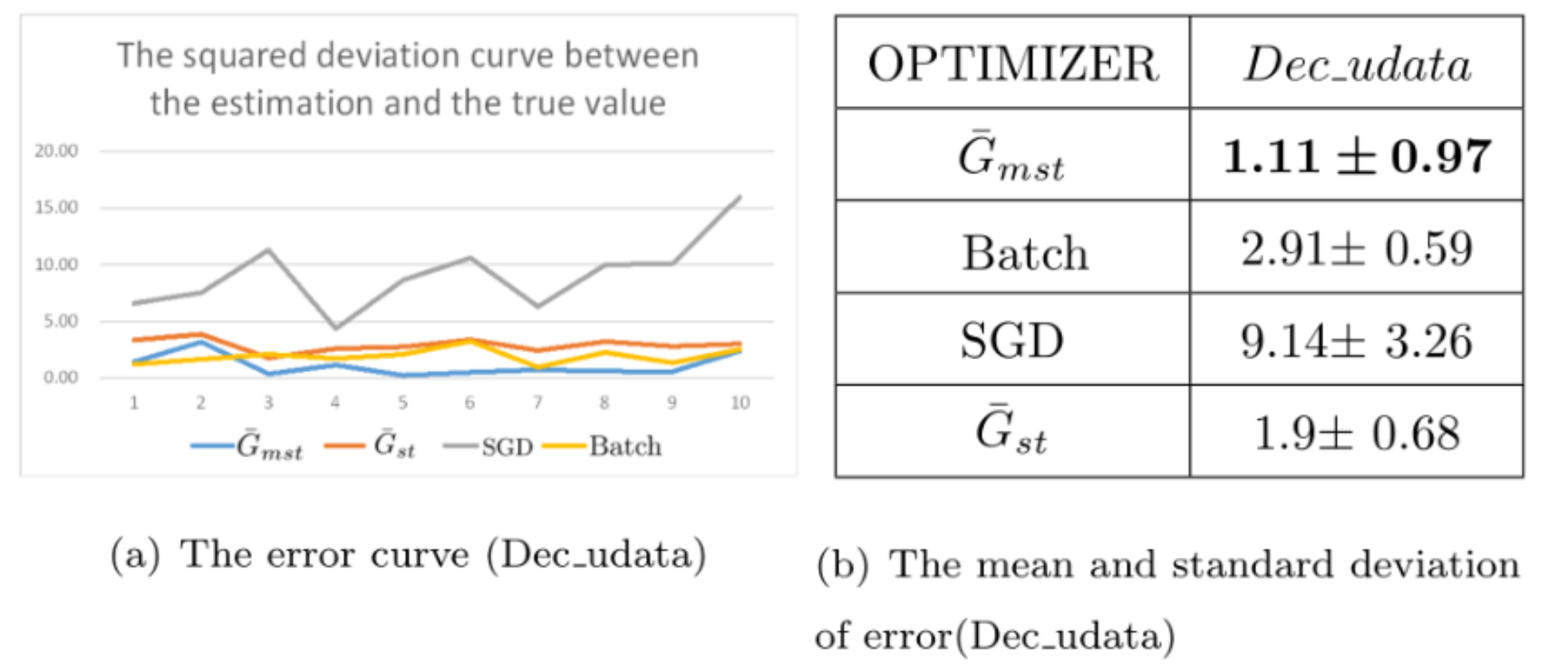}
    \caption{the error curve and its description statistics(Dec$\_$udata)}
    \label{fig:randomdec}
  \end{minipage}%

  \begin{minipage}[t]{\linewidth}
  \centering
    \includegraphics[width=0.8\textwidth]{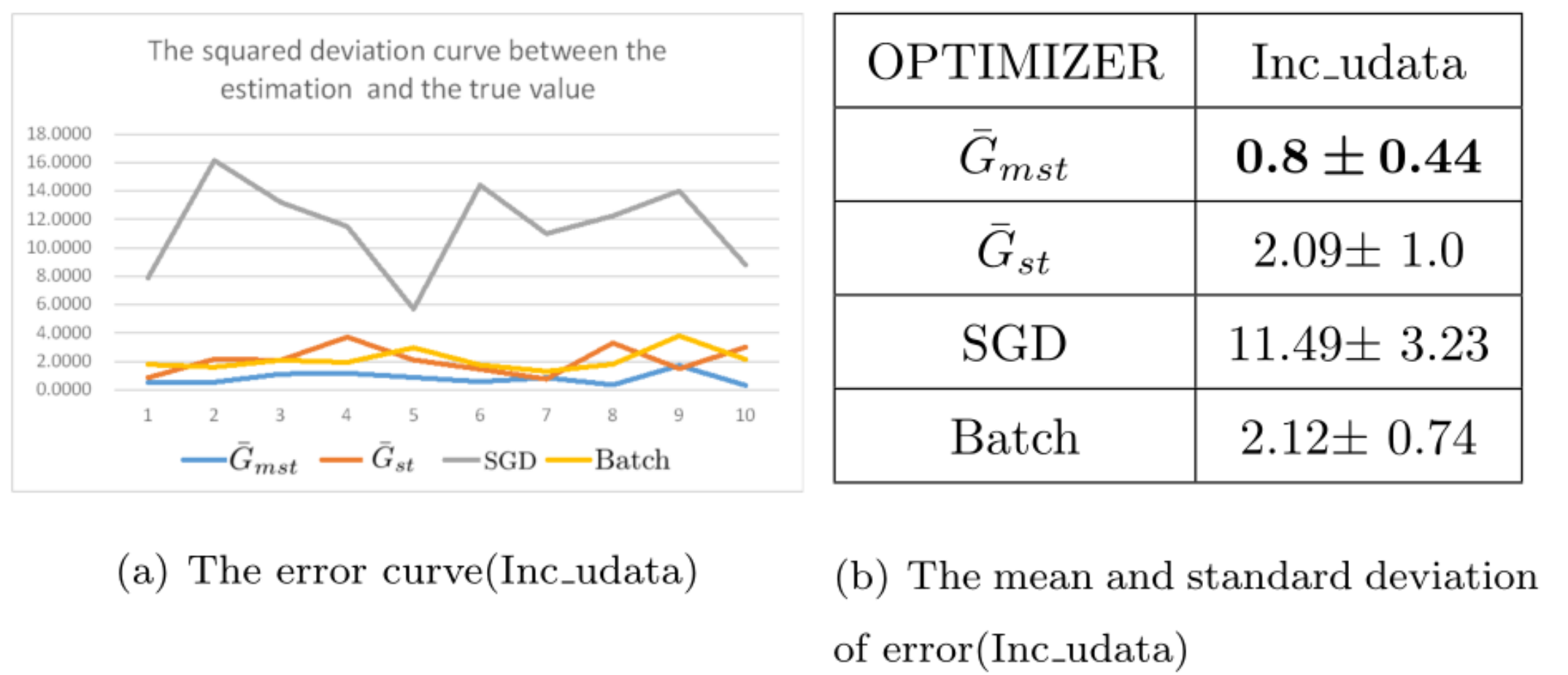}
    \caption{the error curve and its description statistics(Inc$\_$udata)}
    \label{fig:randominc}
  \end{minipage}

\end{figure}

The error in this paper is uniformly expressed by the square of the deviation between the estimator and the true value. Figure \ \ref{fig:randomdec}\  (a) and (b) are the error curve of each estimator and the corresponding error descriptive statistics. As can be seen in Figure 1 (a), the blue curve corresponding to the \ $\bar{G}_{mst}$\  is located at the bottom of all other curves, indicating that the estimated value provided by the \ $\bar{G}_{mst}$\  statistic is the closest to the true value. Figure \ \ref{fig:randomdec}\ (b) shows that the error of \ $\bar{G}_{mst}$\  has the smallest mean and standard deviation, indicating that the estimated value provided by \ $\bar{G}_{mst}$\  is more accurate and more stable.

The previous random data set comes from populations with decreasing mean. The second random data is a set of uniformly distributed random numbers from another \ $10$\  intervals\ $[0,3],[2,3],[2,4],[3,5],[3,6],[4,7],[5,8],[6,9],[8,10],[8,12]$\ (increasing mean), also be formatted into a\ $40\times 10$\ matrix as the data population.

It can be seen from Figure\ \ref{fig:randominc}\  that on the mean increasing data set, the search direction provided by \ $\bar{G}_{mst}$\  is also satisfactory. For the increasing mean data set, in Figure\ \ref{fig:randominc}\  (a), the blue curve corresponding to \ $\bar{G}_{mst}$\   is also located at the bottom of all other curves. In Figure\ \ref{fig:randominc}\  (b), the error of \ $\bar{G}_{mst}$\  also has the smallest mean and standard deviation, so the estimated value provided by \ $\bar{G}_{mst}$\ is more accurate and more stable.

\subsection{Results on normal random data set}
In order to further investigate the performance of the algorithm on other random data sets, this paper generated a set of \ $40\times 10$\  random numbers from the normal population \ $N(\mu,\sigma)$\  as experimental data. According to the different methods of taking the normal population parameters \ $\mu,\sigma$\ ,five different types of data sets are generated: \ random$\_$ndata\ (\ $\mu,\sigma$\ are both uniformly distributed random numbers in the interval of [1,20]), \ MeanD$\_$ndata\ (random data set with decreasing mean), \ MeanI$\_$ndata\ (random data set with increasing mean), \ VarD$\_$ndata\ (random data set with decreasing variance), \ VarI$\_$ndata\ (random data set with increasing variance).

\begin{figure}[htbp]
  \centering
    \includegraphics[width=0.8\textwidth]{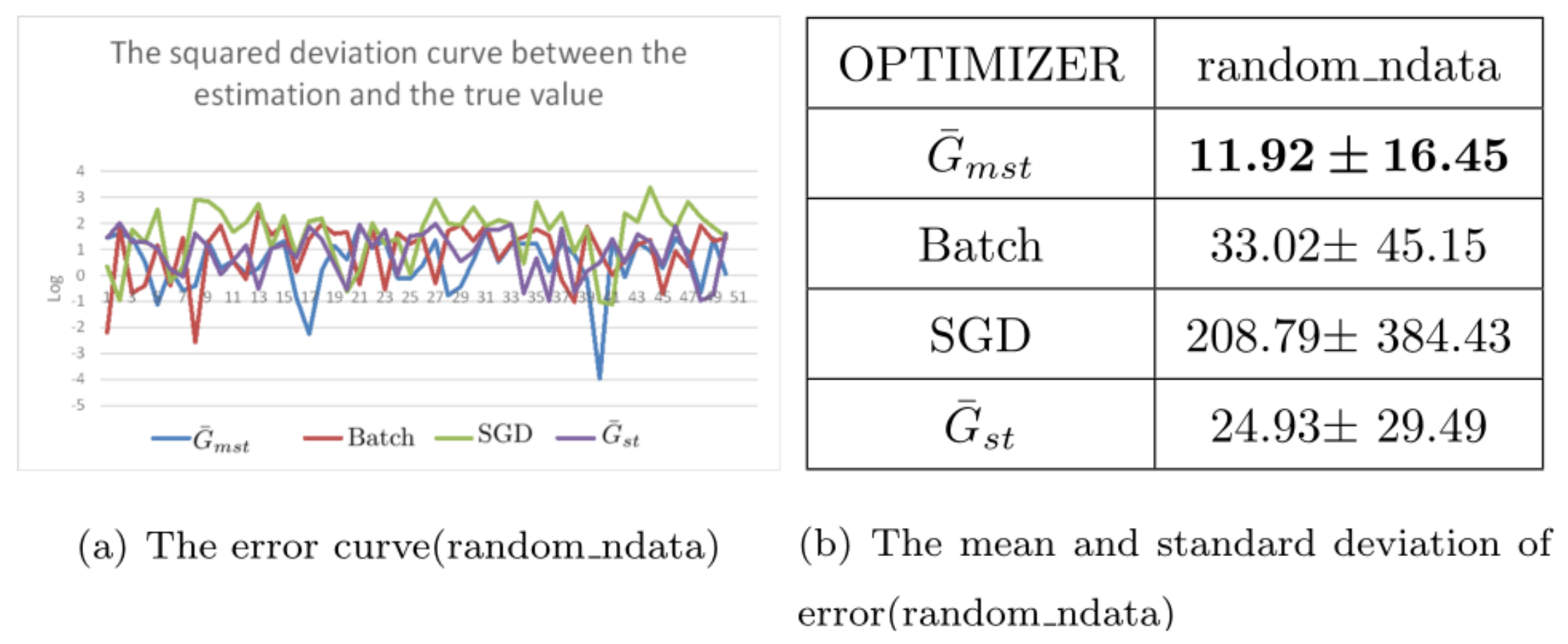}
    \caption{\small{the error curve and its description statistics(random$\_$ndata)}}
    \label{fig:randomn}
\end{figure}

\begin{figure}[htbp]
\centering
  \begin{minipage}[t]{0.45\linewidth}
  \centering
    \includegraphics[width=\textwidth]{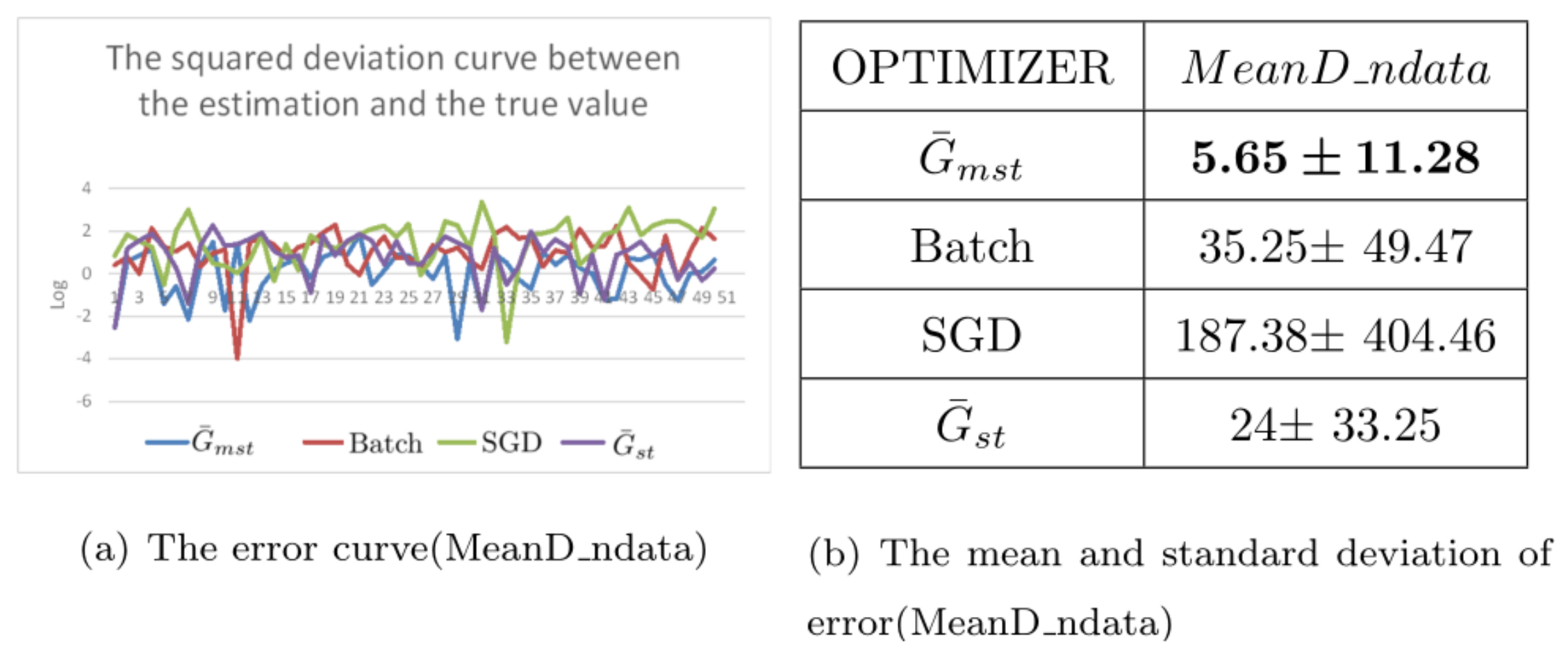}
    \caption{\small{the error curve and its description statistics(MeanD$\_$ndata)}}
    \label{fig:meand}
  \end{minipage}
  \begin{minipage}[t]{0.45\linewidth}
  \centering
    \includegraphics[width=\textwidth]{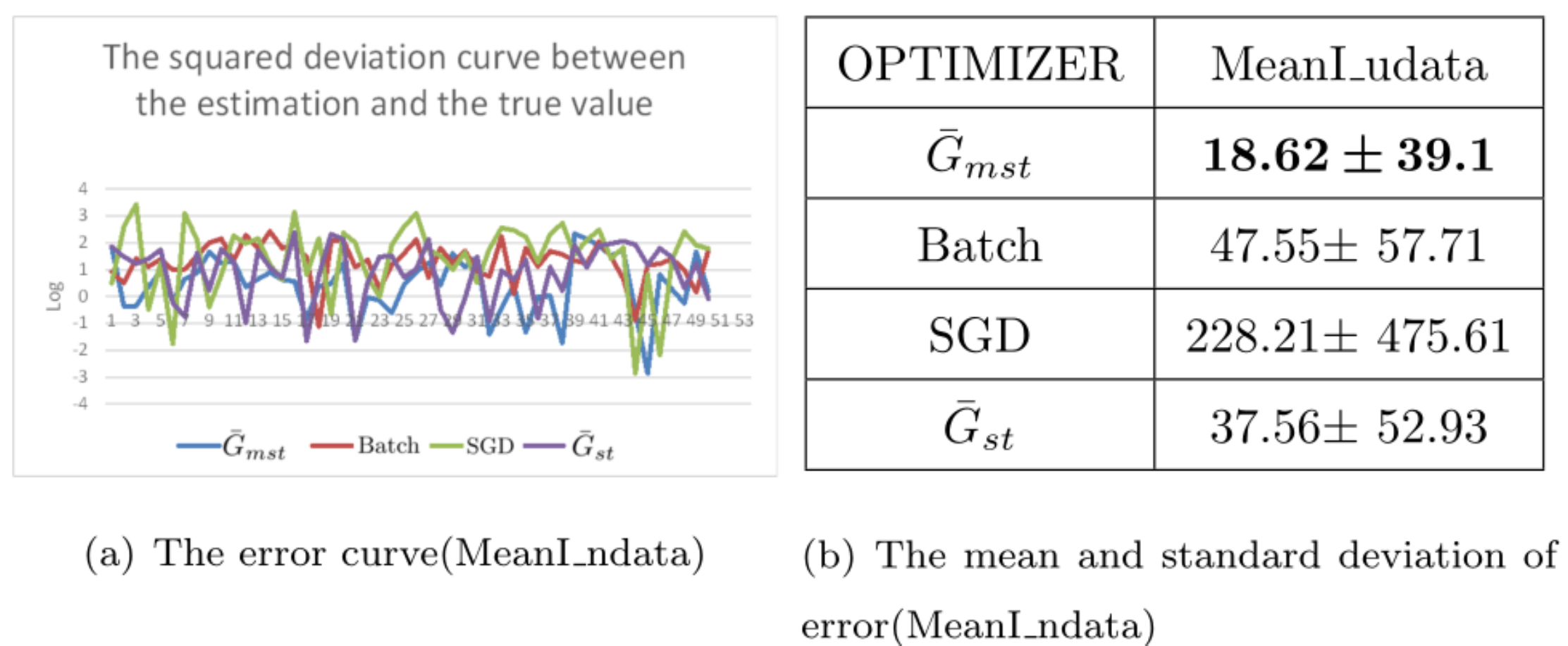}
    \caption{\small{the error curve and its description statistics(MeanI$\_$udata)}}
    \label{fig:meani}
  \end{minipage}

  \begin{minipage}[t]{0.45\linewidth}
  \centering
    \includegraphics[width=\textwidth]{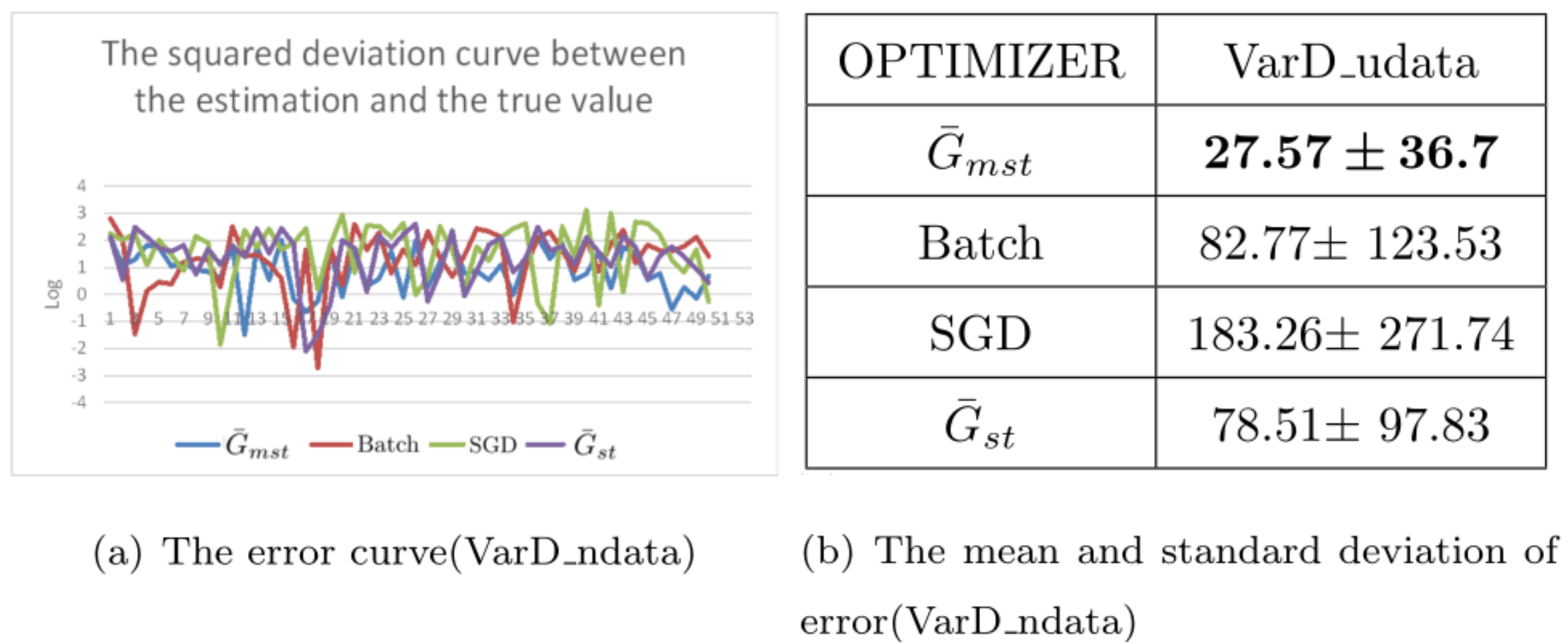}
    \caption{\small{the error curve and its description statistics(VarD$\_$udata)}}
    \label{fig:vard}
  \end{minipage}
  \begin{minipage}[t]{0.45\linewidth}
  \centering
    \includegraphics[width=\textwidth]{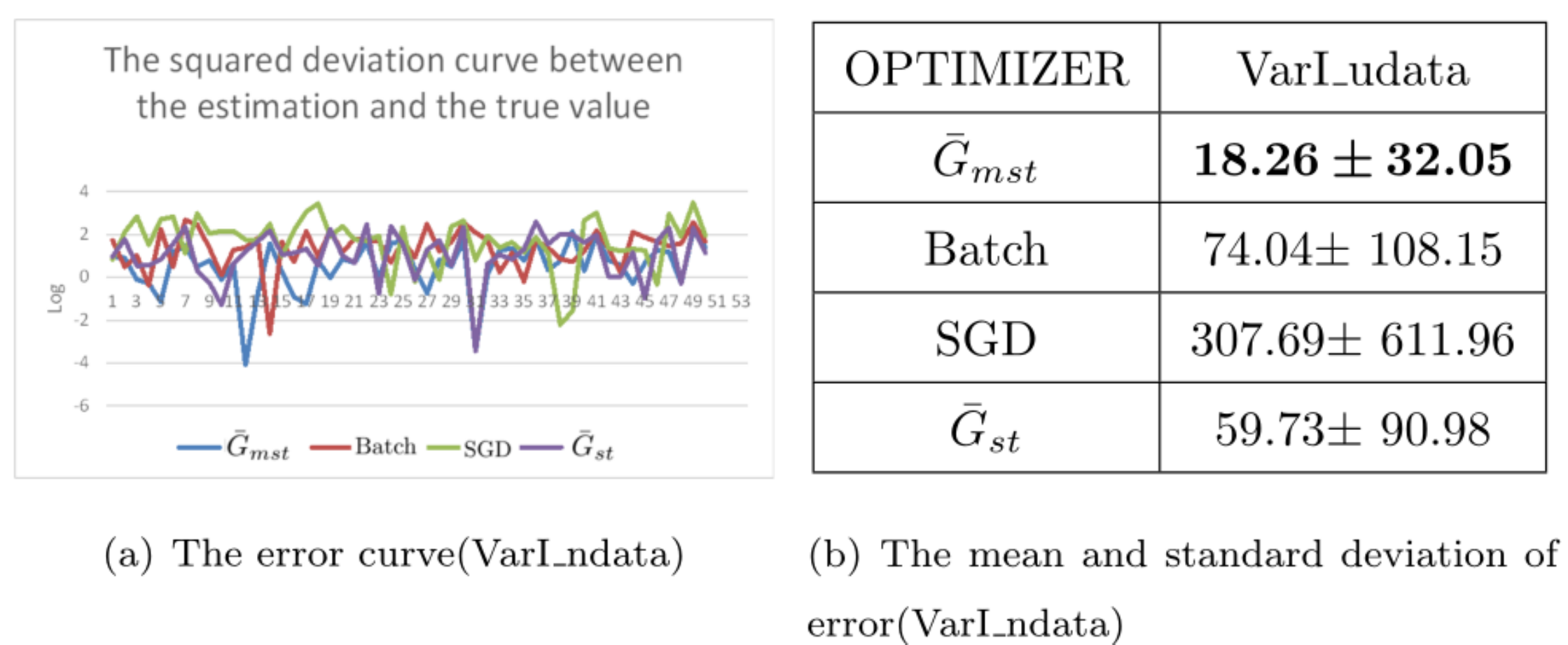}
    \caption{\small{the error curve and its description statistics(VarI$\_$udata)}}
    \label{fig:vari}
  \end{minipage}

\end{figure}

It can be seen from Figure \ \ref{fig:randomn}$\sim$\ref{fig:vari}\  that whether it is a uniformly distributed random normal data set, or a decreasing mean, increasing mean, decreasing variance, and increasing variance data sets, the gradient estimation value generated by the statistic \ $\bar{G}_{mst}$\  mentioned in this article has the smallest error, and its standard deviation is also the smallest.
This shows that the statistic \ $\bar{G}_{mst}$\  can more accurately and stably approximate the gradient mean (true value) of population.

\subsection{Results on the MNIST dataset}
In order to compare the performance differences of different algorithms on the MNIST data set, we use the full gradient method to train a 5-layer forward network with a structure of [784,500, 500,200,10] with a training set of 60,000 scale, with \ $\alpha=0.2$\ (step size),$\lambda=0.001$(weight decay coefficient)\ . After 60 iterations, the network achieves 87.73\% accuracy on the 10,000-scale test set. We recorded the weight w between the first neuron in the output layer and the first neuron in the penultimate layer, forming the gradient information about 60,000 training samples in 60 iterations, forming a gradient matrix with a scale of \ $60,000\times 60$\ . The average value of each column of the gradient matrix is the true gradient direction of the parameter update.

On the \ $60,000\times 60$\  gradient matrix, we calculate the expected \ $E_j$\  and variance \ $V_j$\  of each category \ $j$\  from \ $0\sim 9$\ , then calculate the \ $p_j,q_j$\ of each category according to formulae \ \ref{eq:valuepq}\ . Finally, the value of the \ $\bar{G}_{mst}$\  can be calculated using these parameter values.

In order to enhance comparability, except that SGD is a single sample, the sample sizes of \ $Batch,\bar{G}_{mst},\bar{G}_{st}$\  are all set to 10. Among them, \ $\bar{G}_{mst},\bar{G}_{st}$\  randomly selects a single sample from each of the 10 categories when sampling, and Batch randomly selects 10 samples from 60,000 samples.

\begin{figure}
\centering
\subfigure[$\bar{G}_{mst}$ VS. Pop.]{\includegraphics[width=0.45\textwidth]{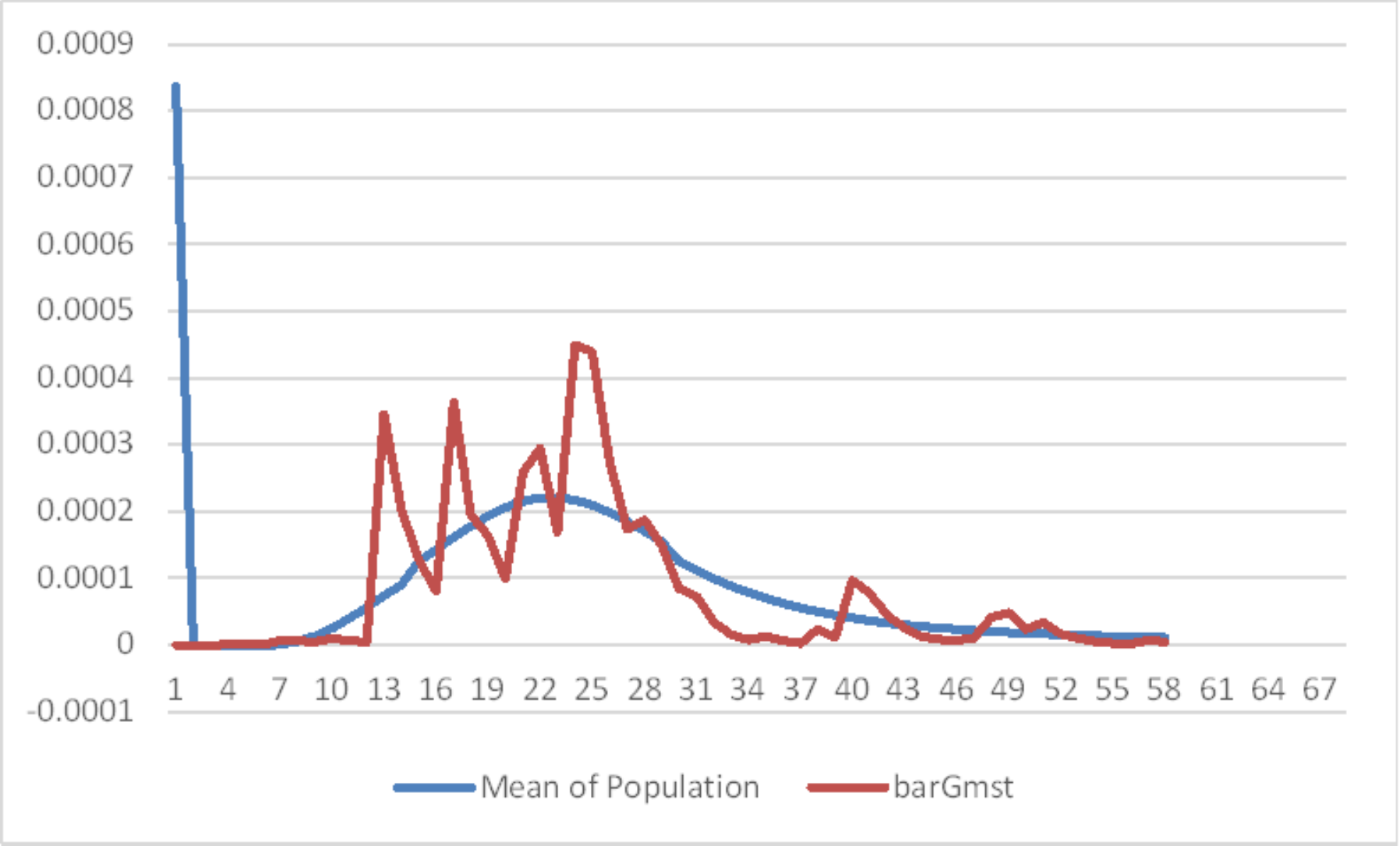}}
\subfigure[$\bar{G}_{st}$ VS. Pop.]{\includegraphics[width=0.45\textwidth]{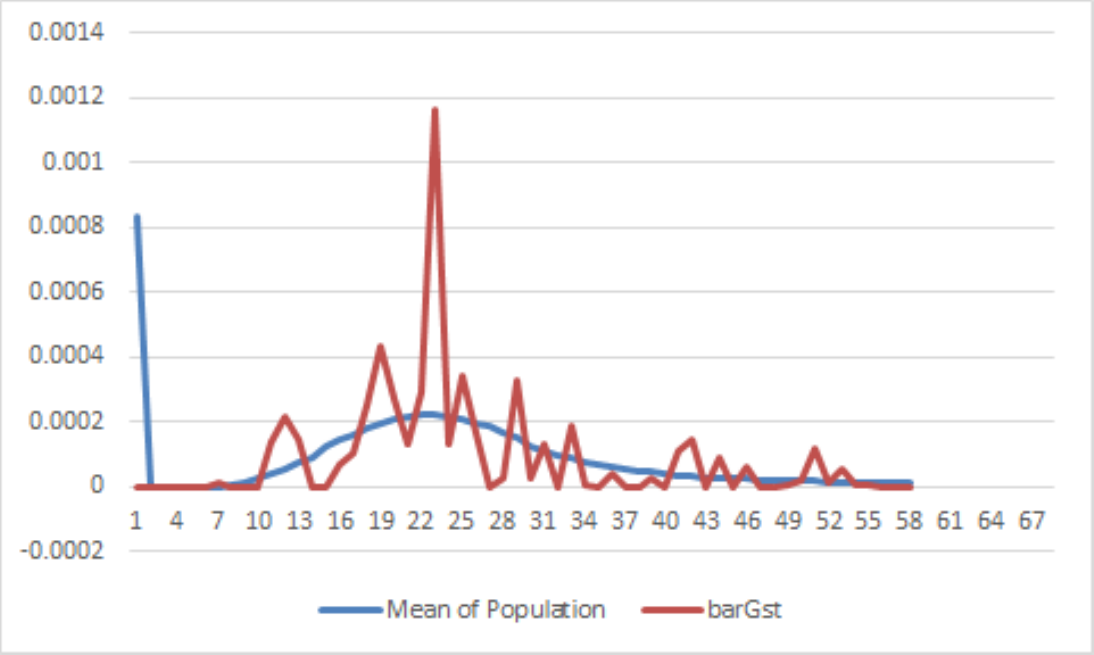}}
\\
\centering
\subfigure[Batch VS. Pop.]{\includegraphics[width=0.45\textwidth]{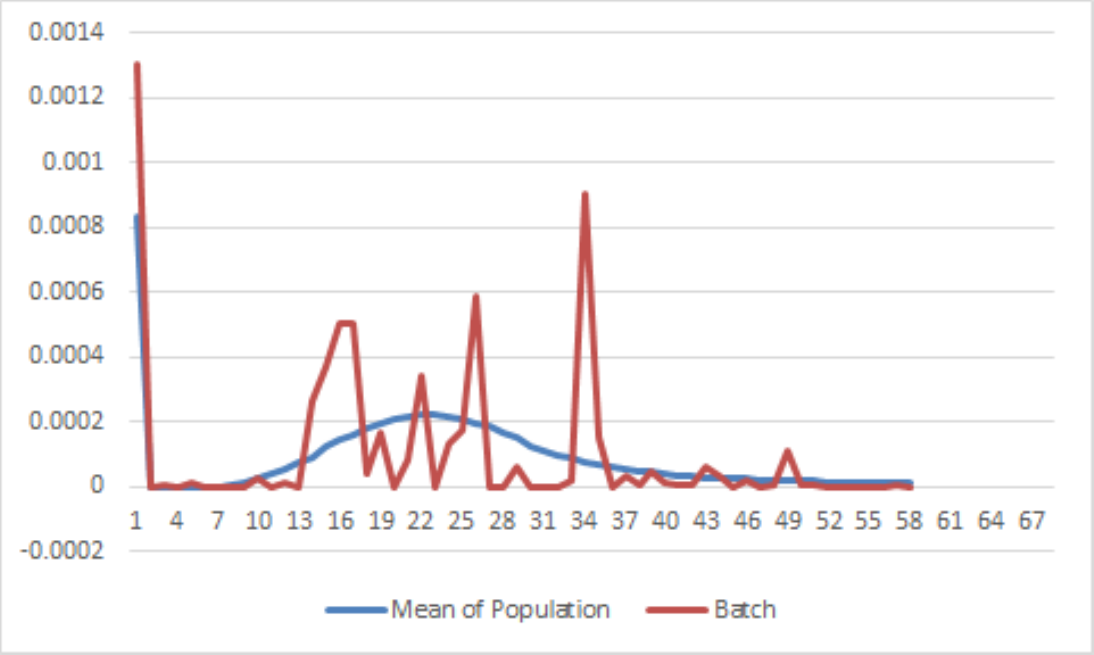}}
\subfigure[SGD VS. Pop.]{\includegraphics[width=0.45\textwidth]{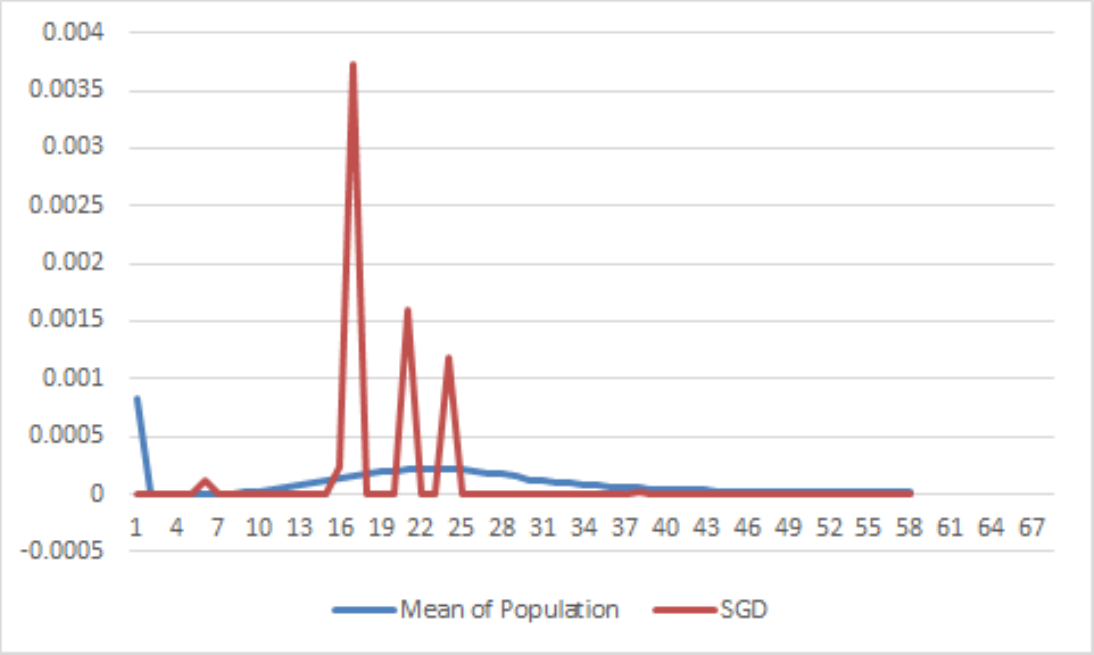}}
\caption{Tracing curves of four different estimators to true values}
\label{fig:mnist}
\end{figure}

The blue curve in Figure \ \ref{fig:mnist}\  represents the true gradient direction(referred to by Pop in Figure \ \ref{fig:mnist}\ ), which is the average gradient sequence generated by the full gradient method after 60 iterations. The red curve is the curve formed by the gradient sequence generated by 60 iterations of different algorithms. The four sub-graphs show the subtle differences in tracking the direction of the blue curve by the gradient curves generated by different methods.

In order to further investigate the accuracy difference of the four methods of \ $Batch,SGD,\bar{G}_{mst},\bar{G}_{st}$\ , we calculate and record the deviation square of the random gradient direction generated by the four methods and the true gradient direction. Thus, each method obtains 60 such deviation squares, and the average and standard deviation of these 60 deviation squares are used to measure the advantages and disadvantages of the algorithm.

\begin{figure}[htbp]
\centering
\includegraphics[width=0.8\textwidth]{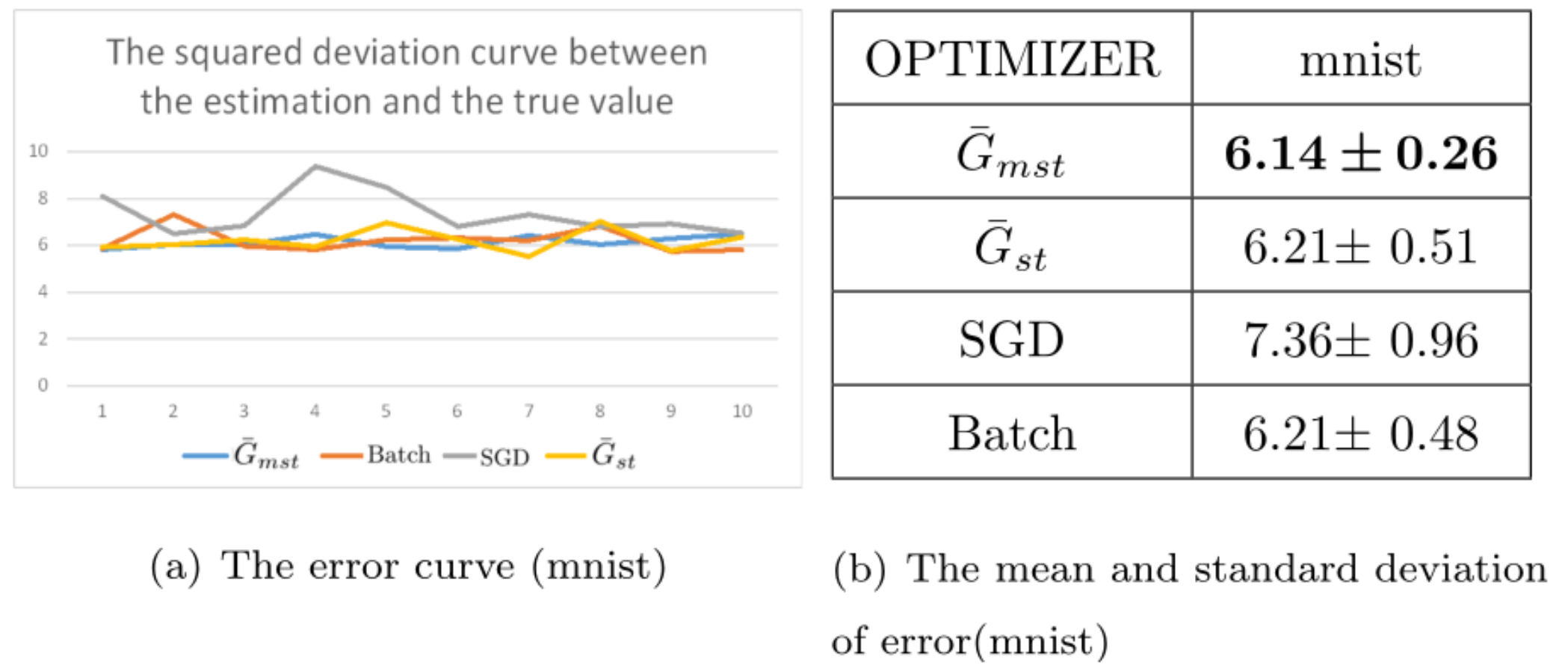}
\caption{\small{the error curve and its description statistics(mnist)}}
\label{fig:mnist2}
\end{figure}

Each algorithm is run repeatedly for 10 times, and the deviation square values generated  are recorded. It can be seen from Figure \ \ref{fig:mnist2}\  that even in the case of a small sampling ratio \ $f=\frac{10}{60000}$\ , the search direction provided by \ $\bar{G}_{mst}$\  is closer to the true value than other methods (the mean and standard deviation of the deviation square are the smallest).

we further compare the performance of the four algorithms \ $Batch,SGD,\bar{G}_{mst},\bar{G}_{st}$\ in optimizing a 5-layer forward network [784,500, 500,200,10]. The data used is still the 60,000-scale MNIST training set and the 10,000-scale MNIST test set. But to be fair, we first implement a grid search procedure for the optimal hyperparameters in the range of alpha=[0.01,1,0.001], lambda=[0.001,0.0001],then examine the training and test accuracy differences of the four algorithms  after 1, 2, 3, 4, 5, 6, 7, 8, 9, and 10 thousand iterations under their respective optimal hyperparameters.

Table\ \ref{tab:traintestaccu}\  shows the training and testing accuracy achieved by the four algorithms \ $Batch,SGD,\bar{G}_{mst},\bar{G}_{st}$\  under their respective optimal hyperparameters. As can be seen from Table \ \ref{tab:traintestaccu}\ , except for 7 and 9 thousand iterations, the training and the test accuracy of \ $\bar{G}_{mst}$\  are the best, outperform than the other four algorithms.
\begin{table}[htbp]
  \centering
  \caption{Add caption}
  \scalebox{0.6}{
    \begin{tabular}{|c|c|c|c|c|c|c|c|c|}
    \hline
    Iterations & \multicolumn{2}{c|}{SGD(\%)} & \multicolumn{2}{c|}{$\bar{G}_{mst}$(\%)} & \multicolumn{2}{c|}{Batch(\%)} & \multicolumn{2}{c|}{$\bar{G}_{st}$(\%)} \\
    \cline{2-9}
        ($10^3$)  & test accu & train accu & test accu & train accu & test accu & train accu & test accu & train accu \\
        \hline
    1(SGD:$\times$20) & 91.48 & 91.39 & \textbf{94.46} & \textbf{94.56} & 93.31 & 93.38 & 94.12 & 94.28 \\
    2(SGD:$\times$20) & 92.87 & 92.94 & \textbf{96.05} & \textbf{96.37} & 95.52 & 95.88 & 95.92 & 96.18 \\
    3(SGD:$\times$20) & 93.83 & 94.08 & \textbf{96.53} & \textbf{96.93} & 96.35 & 96.86 & 96.17 & 96.42 \\
    4(SGD:$\times$20) & 94.57 & 94.76 & \textbf{96.74} & \textbf{97.39} & 96.61 & 97.25 & 96.44 & 96.78 \\
    5(SGD:$\times$20) & 94.7  & 94.79 & \textbf{97.17} & \textbf{97.98} & 97.12 & 97.81 & 96.65 & 97.18 \\
    6(SGD:$\times$20) & 95.26 & 95.8  & \textbf{97.3}  & \textbf{98.11} & 97.16 & 97.91 & 96.94 & 97.39 \\
    7(SGD:$\times$20) & 95.05 & 95.7  & 97.13 & \textbf{98.19} & \textbf{97.23} & 97.89 & 96.64 & 97.22 \\
    8(SGD:$\times$20) & 95.55 & 95.82 & \textbf{97.65} & \textbf{98.4}  & 97.43 & 98.33 & 96.62 & 97.09 \\
    9(SGD:$\times$20) & 95.66 & 96.24 & 97.41 & 98.41 & \textbf{97.72} & \textbf{98.49} & 96.92 & 97.42 \\
    10(SGD:$\times$20) & 95.44 & 95.9  & \textbf{97.63} & \textbf{98.81} & 97.4  & 98.45 & 97.31 & 97.81 \\
    \hline
    \end{tabular}%
    }
  \label{tab:traintestaccu}%
\end{table}%

\section{Related work}

From the update formula \ $\upsilon_t=\gamma\upsilon_{t-1}+\eta\nabla_{\theta} J(\theta),\theta=\theta-\upsilon_t$\  of the Momentum optimization\ \cite{Qian1999}, the gradient direction required for parameter update in the Momentum optimization is obtained by the weighted summation of the current gradient \ $\nabla_{\theta} J(\theta)$ and the historical gradient information stored in \ $\upsilon_{t-1}$\  with the weights \ $\eta,\gamma$\ . Here,  \ $\eta,\gamma$\  are similar to that of the parameters q and p, The role of \ $\upsilon_{t}$\  is equivalent to the auxiliary variable G in this article. However, the Momentum optimization does not discuss the unbiasedness of \ $\upsilon_{t}$\ . The results of this paper show that if \ $\eta,\gamma$\ satisfy the \ $q,p$\  conditions in formula \ \ref{eq:valuepq}\ , the search direction provided by the Momentum optimization satisfies unbiasedness.

The widely popular Adam algorithm\ \cite{Kingma2015}\  for training deep models,its gradient update formula is \ $m_t=\beta_1m_{t-1}+(1-\beta_1)g_t,\hat{m}_t=\frac{m_t}{1-\beta_1}$. The parameters \ $\beta_1,(1-\beta_1)$\ therein are respectively equivalent to the \ $p,q$\  parameters in this article. Obviously, according to the results of this article, one of the prerequisites for the effectiveness of Adam's coefficients \ $\beta_1,(1-\beta_1)$\ of \ $m_{t-1},g_t$\  is to ensure that the mean value of the gradients before and after the iterations are equal. Generally, if without additional restrictions, the equal-mean properties of the gradient before and after the iteration are generally not satisfied. Therefore, the author of the Adam algorithm made a so-called unbiased correction \ $\hat{m}_t=\frac{m_t}{1-\beta_1}$\  on \ $m_t$\ . Obviously, this is an empirical correction formulae, the theoretical basis behind the correction is not fully understood by the authors. In fact, the revised estimator must be a biased estimator, which is contrary to the original intention of the proponent.

The SAG algorithm\ \cite{Roux2012}\ that claims to achieve a linear convergence rate under the condition of strong convexity of the optimization target, the gradient \ $f'_i(x^k)$\  of the current random sample has a proportion of \ $\frac{1}{n}$\  in the parameter update direction (where n is the total number of training samples), and the remaining ratios are averaged Divided into the gradient of different random samples in different iterations. Here the weight of \ $\frac{1}{n}$\  is equivalent to the parameter q in this paper.  However, SAG roughly assumes that changes in network parameters, such as changes from \ $x^{k-1}$\  to \ $x^{k}$\ , or other forms of change, will not cause fluctuations in the gradient expectations, that is, \ $E(f'_i(x^k))=E(f'_i(x^{k-1}))$. Obviously, the theoretical convergence results obtained on the basis of such assumptions are not firm. Other algorithms such as SAGA\ \cite{Defazio2014}\  and SVRG\ \cite{Johnson2013}\  also explicitly or implicitly acknowledge similar assumptions.

Other variance reduction methods that perform k-step averaging on historical trajectories are equivalent to taking p and q in this article as \ $\frac{k-1}{k},\frac{1}{k}$\ respectively. However, if this approach does not have an additional strategy to limit the gradient to ensure equal gradient mean between different iterations, then the unbiasedness of this approach cannot be satisfied, and the effect of the algorithm will be difficult to guarantee.

\end{document}